\documentclass{article}

\usepackage{arxiv}

\usepackage[utf8]{inputenc} 
\usepackage[T1]{fontenc}    
\usepackage{hyperref}       
\usepackage{url}            
\usepackage{booktabs}       
\usepackage{amsfonts}       
\usepackage{nicefrac}       
\usepackage{microtype}      

\usepackage{lipsum}         
\usepackage{graphicx}
\usepackage[square,numbers]{natbib}
\usepackage{doi}

\usepackage{xcolor}         
\usepackage{amsmath}
\usepackage{amsthm}
\usepackage{graphicx}
\usepackage{bm}
\usepackage[ruled]{algorithm2e}
\usepackage{xspace}
\usepackage{caption}
\usepackage{subcaption}
\usepackage{makecell}
\usepackage{cleveref}       
\newcommand{\myparagraph}[1]{\noindent\textbf{#1}}

\DeclareMathOperator*{\argmin}{arg\,min}
\newtheorem{theorem}{Theorem}
\SetKwComment{Comment}{$\triangleright$\ }{}
\newcommand{\Ls}{\mathcal{L}}
\newcommand{\Lz}{\overline{\mathcal{L}}_z}

\title{M²FGB: A Min-Max Gradient Boosting Framework for Subgroup Fairness}

\newif\ifuniqueAffiliation
\uniqueAffiliationtrue

\ifuniqueAffiliation 
\author{%
Jansen S. B. Pereira\thanks{The first two authors contributed equally to this work.}\\
Instituto de Computação \\
Universidade Estadual de Campinas \\
\texttt{j252477@dac.unicamp.br}
\And
Giovani Valdrighi\footnotemark[1] \\
Instituto de Computação \\
Universidade Estadual de Campinas \\
\texttt{giovani.valdrighi@ic.unicamp.br}
\And
Marcos Medeiros Raimundo \\
Instituto de Computação \\
Universidade Estadual de Campinas \\
\texttt{mraimundo@ic.unicamp.br}
}
\else
\usepackage{authblk}

\newbox{\orcid}\sbox{\orcid}{\includegraphics[scale=0.06]{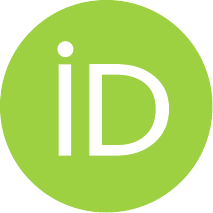}}
\author[1]{%
Jansen S. B. Pereira%
\thanks{\texttt{j252477@dac.unicamp.br}}%
}
\author[1]{%
Giovani Valdrighi%
\thanks{\texttt{giovani.valdrighi@ic.unicamp.br}}%
}
\author[1]{%
Marcos Medeiros Raimundo%
\thanks{\texttt{mraimundo@ic.unicamp.br}}%
}
\affil[1]{Instituto de Computação, Universidade Estadual de Campinas, Campinas, SP, Brasil}
\fi

\renewcommand{\shorttitle}{\textit{arXiv} Template}

\hypersetup{
pdftitle={A template for the arxiv style},
pdfsubject={q-bio.NC, q-bio.QM},
pdfauthor={David S.~Hippocampus, Elias D.~Striatum},
pdfkeywords={First keyword, Second keyword, More},
}

\begin{document}
\maketitle

\begin{abstract}
In recent years, fairness in machine learning has emerged as a critical concern to ensure that developed and deployed predictive models do not have disadvantageous predictions for marginalized groups. It is essential to mitigate discrimination against individuals based on protected attributes such as gender and race. In this work, we consider applying subgroup justice concepts to gradient-boosting machines designed for supervised learning problems. Our approach expanded gradient-boosting methodologies to explore a broader range of objective functions, which combines conventional losses such as the ones from classification and regression and a min-max fairness term. We study relevant theoretical properties of the solution of the min-max optimization problem. The optimization process explored the primal-dual problems at each boosting round. This generic framework can be adapted to diverse fairness concepts. The proposed min-max primal-dual gradient boosting algorithm was theoretically shown to converge under mild conditions and empirically shown to be a powerful and flexible approach to address binary and subgroup fairness.
\end{abstract}


\section{Introduction}

Fairness has been a recurring study topic in recent years due to the numerous examples of biased decision-making algorithms identified in high-impact applications, such as crime recidivism~\cite{angwin2016machine}, recruiting process~\cite{dastin2018amazon}, and others. \textit{Group justice} is the most utilized definition of fairness, which demands the independence of the predictions and a sensitive attribute, such as gender or race~\cite{mehrabi2021survey}. This definition can also be interpreted as requiring no disparity in the allocation of resources among privileged and unprivileged groups~\cite{mehrabi2021survey} or no disparity in error rates between groups~\cite{hardt2016equality}. In that sense, most fairness algorithms consider minimizing a prediction loss with a constraint on the difference of utility measures. However, this may achieve fairness by only reducing the quality of the predictions for a group, which is not feasible in some applications. Based on Rawlsian ethics~\cite{rawls2001justice}, \textbf{min-max fairness} aims to increase the well-being of the marginalized group. This concept becomes more complex when considering multiple demographic groups~\cite{kearns2018preventing}.

Considering an example in the scenario of health diagnostics, where fairness is evaluated with combined attributes of race (white and Black) and gender (man and woman) in four subgroups. Fairness principles based on disparity have the objective of reducing the gap between the highest and lowest group-level errors. If there is an initial solution where the errors are $e_1 < e_2 < e_3 < e_4; e_i \in \mathbb{R}$, the equalized loss metric is $\max_{i \neq j} |e_j - e_i| = e_4 - e_1$. However, another classifier with losses $e_1 + \epsilon < e_2 < e_3 < e_4, \epsilon > 0 \in \mathbb{R}$, would have the equalized loss metric equal to $\max_{i \neq j} |e_j - e_i| = e_4 - e_1 - \epsilon < e_4 - e_1 $. Although this new classifier has equal or worse performance for all individuals, it is evaluated as better by the equalized loss metric. This was obtained by harming a subgroup without improving the condition of another one. In the min-max setting, the second classifier will not be evaluated as an improvement from the initial one. This is a common situation where $e_4$ might be the best loss possible atainable for the worst-performing group under the available data.

Previous studies have already considered min-max fairness algorithms. \citet{martinez2020minimax} presented an initial work using this definition, which considered a multi-objective optimization problem where each group-level error was an objective. The proposed algorithm searches for a weighted sum of the objectives that results in a solution to the original min-max problem. \citet{diana2021minimax} proposed a two-player game formulation where one player minimizes the group-level losses at each round while the other finds a maximum weighted sum of the losses. However, both algorithms encompass fitting multiple logistic models, which can have high computational costs, particularly with large datasets. The computing and accuracy costs of such fairness solutions can impede decision-makers from utilizing them. Considering the scenario of tabular data, a common data type for social domains, gradient-boosting algorithms have excellent results in accuracy and computing cost~\cite{shwartz2022tabular}. Despite these algorithms being already considered for fairness~\cite{ravichandran2020fairxgboost, cruz2023fairgbm}, there is still no adaptation considering min-max fairness.

In this setting of min-max fairness, we propose \textbf{M²FGB}, an algorithm that leverages the competitive accuracy and efficiency of gradient-boosting algorithms to maximize the worst-group (WG) utility, i.e., min-max fairness. By considering the Lagrangian function, the algorithm performs primal-dual optimization within the gradient-boosting rounds by first performing a gradient ascent on dual variables and then a gradient descent to build the ensemble model. Although our algorithm asks for differentiable measures of fairness, we present diverse proxy functions for different fairness criteria. The empirical results showed that optimizing for the WG utility with the proxy differentiable fairness measure also improves upon the non differentiable measure of interest. M²FGB is versatile, working with different loss functions and fairness metrics. It can also optimize fairness metrics defined for sensitive attributes with more than two categories. In summary, the main contributions of this paper are:

\begin{itemize}
    \item M²FGB, a novel and versatile framework for optimizing min-max fairness through gradient-boosting;
    \item The derivation of our method at three different fairness criteria: equalized loss (for classification and regression), true positive rate (related to equal opportunity difference), and positive rate (related to statistical parity difference);
    \item The validation of our framework in different settings and datasets, including German Credit~\cite{german}, COMPAS~\cite{angwin2016machine}, ENEM~\cite{alghamdi2022beyond}, ACSIncome~\cite{retiring2021ding}. M²FGB showed a competitive performance with compared fairness algorithms;
    \item An open-source implementation of M²FGB available at \url{https://github.com/hiaac-finance/m2fgb}.
\end{itemize}

\section{Preliminaries and Problem Statement}

We consider access to a dataset $(x_i,y_i)^{n}_{i=1}$ with $n$ observations, in which $x_i \in \mathbb{R}^d$ corresponds to the input variables and $y_i$ represents the dependent variable. $y_i \in \{0, 1\}$ if we consider a classification task or $y_i \in \mathbb{R}$ if it is a regression task, but the following discussions are valid for both. We also have access to a categorical sensitive attribute $z_i \in \mathcal{Z}$ for each observation that indicates belonging to a specific demographic group, for example $z_i \in \{\text{male, female, non-binary}\}$ or $z_i \in \{\text{(male, white), (male, Black), (female, white), (female, Black), (non-binary, white), (non-binary, Black)}\}$. The initial objective is to reconstruct the unknown mapping $x \overset{\mathcal{F}}{\longrightarrow} y$ using an estimate $f: \mathbb{R}^d \to \mathbb{R}$.

\subsection{Gradient-Boosting}

 \textit{Gradient-Boosting Algorithms} are ensemble methods that create a strong predictor that minimizes a loss function $\mathcal{L}$ by iteratively adding new weak models to the predictor. The weak model is a step in a functional gradient descent at each boosting round. In more detail, at boosting round $t$ with a predictor $f^{(t-1)} : \mathbb{R}^d \to \mathbb{R}$, the gradient-boosting algorithm adds a new weak learner $h^{(t)} : \mathbb{R}^d \to \mathbb{R}$ from an hypothesis space $\mathcal{H}$ to approximate the gradient of the loss $\mathcal{L}$ w.r.t. the predictions $f^{(t-1)}(x)$. All of the gradient-boosting methods -- from seminal~\cite{friedman2001greedy} to quadratic~\cite{chen2016xgboost}, efficient~\cite{guolin2017lgbm}, and categorical~\cite{prokhorenkova2018catboost} adaptations -- are grounded on the same basis. We have three main components: (1) the loss function, which evaluates how well the model fits the observed data; (2) the weak learner, which makes the predictions; and (3) the additive model, a sequential approach to adding models. This framework has flexibility in the loss utilized, which can be, for example, binary cross-entropy or mean squared error. The model updates are as follows:

\begin{equation*}
\begin{split}
 f^{(t)} \gets  f^{(t-1)} + &\argmin_{h^{(t)} \in \mathcal{H}} \mathcal{L}(y,f^{(t-1)} + h^{(t)}) \\
    \hat h^{(t)} \approx - \gamma \dfrac{\partial\mathcal{L}(y,f^{(t-1)})}{\partial f^{(t-1)}} &\Longrightarrow \; f^{(t)} \approx f^{(t-1)} - \gamma \times \dfrac{\partial\mathcal{L}(y,f^{(t-1))}}{\partial f^{(t-1)}} 
\end{split}
\end{equation*}

This iterative update guarantees that $\mathcal{L}(y, f^{(t)}) \leq \mathcal{L}(y, f^{(t-1)})$ when the step size (learning rate) $\gamma$ is small enough. See \cite{friedman2001greedy} for a more in-depth analysis of the algorithm. The most usual application of gradient boosting machines utilizes decision trees as weak learners, with many techniques developed to decrease the computational cost of calculating the approximation. In particular, LGBM~\cite{guolin2017lgbm} implementation achieves the best results in different scenarios.

\subsection{Fairness in Machine Learning}

Machine learning models systematically assigning favorable labels to certain \textit{groups} of the population is unacceptable and unjust. This notion is called \textit{group justice}, and the protected variables are utilized to define population groups~\cite{verma2018fairness}. Protected variables are usually defined in laws or regulations and consider who has historically been more likely to receive favorable labels~\cite{varshney2022trustworthy}. A different view on fairness, called \textit{individual justice}, proposes that ``similar'' individuals should receive ``similar'' model predictions~\cite{varshney2022trustworthy}. \textit{Subgroup justice}~\cite{kearns2018preventing} also divides the population into groups but considers the intersection of protected attributes\footnote{For example, a model may be fair in terms of gender and race separately, but not at the intersection of gender and race. For instance, the model may not be impartial in assigning favorable labels to white men compared to Black women.}. In this work, we tackle group and subgroup fairness.

\myparagraph{Fairness Metrics} When considering multiple groups (subgroups), disparity-based fairness metrics can be formulated as the difference between maximum and minimum group-level metrics~\cite{ghosh2021characterizing}, with the following general formulation:

\begin{equation*}
    \mathcal{D} = \max_{z' \in \mathcal{Z}}\mathbb{E}[\mu(f(x)) | Z = z'] - \min_{z' \in \mathcal{Z}}\mathbb{E}[\mu(f(x)) | z = z']
\end{equation*}
where $\mu$ is a utility function that can be the identity function in the setting of \textit{Statistical Parity}~\cite{corbett2017algorithmic}, it can be $(f(x) | y =1)$ in the setting of \textit{Equality of Opportunity}~\cite{hardt2016equality} or a measure of accuracy such as $\bm{1}_{[f(x) = y]}$ in the setting of \textit{Overall Accuracy Equality}~\cite{berk2021fairness}. Commonly, fairness metrics are not differentiable, which be an obstacle in the development of optimization algorithms.

\myparagraph{Rawlsian Based Min-max Fairness} One of the main drawbacks of previous fairness definitions/metrics is the intrinsic assumption that closer metrics between groups are necessarily better in a broader context. In some cases, reducing the rate of positives or false positives does not necessarily improve the `unprivileged' group but only makes the `privileged' group worse~\cite{martinez2020minimax}. To deal with such a challenge, we rely on understanding Rawls' ethics on `Justice as fairness: A restatement'~\cite{rawls2001justice} that involves minimizing the maximal error/loss among groups (or maximizing the minimal utility)~\cite{wan2023processing}. One example of common metric in this setting is `\textbf{worst-group accuracy}', defined as $\min_{z' \in \mathcal{Z}} \mathbb{E}[\bm{1}_{[f(X) = Y]} | Z = z']$. Min-max fairness metrics can be generalized as $\max_{z' \in \mathcal{Z}} \mathbb{E}[\mu(f(X)) | Z = z']$ where $\mu$ is a measure of harm. We opt to utilize the `maximum harm among groups' to be in accordance to conventional optimization formulations.

In this work, we consider learning an estimator $f$ that optimizes Ralwsian's fairness metrics, i.e., min-max fairness. This problem was previously studied by \citet{martinez2020minimax}. This problem has been previously formulated as an optimization problem by using a loss function $\overline{\mathcal{L}}_z(\bm{y},f(\bm{x}))$ that is defined for each group $z \in \mathcal{Z}$ and represents the loss for this specific population. This may be obtained by calculating a conventional loss within only the samples from group $z$. Then, the objective is to \textbf{minimize the maximum group-level loss}. The problem is stated as follows:

\begin{align}
\label{eq:minimax}
P: \quad \min_{f} \quad \max_{z \in \mathcal{Z}} \overline{\mathcal{L}}_z(\bm{y},f(\bm{x}))
\end{align}

This formulation is flexible as different fairness metrics can be formulated in a loss, including for classification and regression tasks as we discuss on Sec.~\ref{sec:other_fairness_metrics}, and is able to handle an arbitrary number of groups. It is important to know that a min-max solution in convex settings will never hurt the performance of a `privileged' group if the `unprivileged' group has no improvement. This is called \textbf{no unnecessary harm}~\cite{martinez2020minimax}.

\section{Related Works}
\label{sec:related_works}

Approaches to bias mitigation in machine learning can be classified into three main categories: pre-processing, in-processing, and post-processing~\cite{varshney2022trustworthy}. \textit{Pre-Processing} methods~\cite{zhang2022review} aims to modify the dataset to mitigate pre-existing biases before model training.
\textit{In-Processing} methods~\cite{wan2023processing} integrate bias reduction strategies during model training. This can be done, for example, by adjusting the loss function to penalize biased predictions. This permits greater flexibility and the ability to obtain algorithms at different points in the fairness-accuracy trade-off. \textit{Post-Processing}~\cite{gouverneurfairness} methods are applied after the model has been trained, which can alter predictions for certain groups or recalibrate output probabilities. Pre-processing and post-processing can be combined with any in-processing algorithm, and for that reason, our focus is on comparing our methodology with other in-processing algorithms.

Initial approaches to in-processing algorithms were predominantly based on minimizing predictor errors with fairness constraints. Constraints were incorporated directly as the disparity measure~\cite{kamishima2012fairness, agarwal2018reductions}, constraining the correlation between the outcome and prediction~\cite{zafar2019fairness}, and with convex surrogates~\cite{wu2018fairness}. 
\citet{kearns2018preventing} introduced subgroup fairness, a middle category between group and individual fairness, which consider the intersectionality of multiple protected attributes. Subsequent works already were designed for the scenario where the sensitive attribute is not binary~\cite{zafar2019fairness, tarzanagh2023fairness, foulds2020intersectional}, for example considering strong demographic parity in classification~\cite{pmlr-v115-jiang20a, barata2021fair} or using random classifiers~\cite{shui2022learning}. Yet, despite these methodologies handling multiple groups, they do not optimize for the min-max criterion.

The area of fairness in machine learning has made advances in min-max fairness in recent years. \citet{martinez2020minimax} presented an initial work that introduced Ralwsian's ethics by considering the loss in each group as an objective to optimize. By using multi-objective optimization, they presented an algorithm capable of identifying the Pareto frontier of solutions. This work was followed by improvements, such as considering improving ROC-AUC among groups~\cite{yang2023minimax}, or unknown demographic~\cite{lahoti2020fairness, martinez2021blind, hashimoto2018fairness}, which tries to maximize the utility of any subgroup of the samples. \citet{diana2021minimax} provided further improvements on this topic, designing an algorithm as a two-player game, where at each iteration, a player fits a model and the other reweights the groups based on the loss. Martinez et al. and Diana et al. algorithms are composed of iterations that train each one a logistic regression or a neural network, which can be really time-consuming and suffer on large datasets. Although our proposal has similarities to these previous two papers, it offers a more efficient algorithm.

Algorithms for learning fair ensemble models have already been studied. Ensembles of models from a Pareto frontier of fairness and accuracy provided good results~\cite {guardieiro2023enforcing}. Similarly to our approach, AdaFair~\cite{iosifidis2019adafair} calculates samples' weights at each boosting round to maximize fairness. Gradient-boosting has already been considered in fairness problems in the individual fairness setting~\cite{vargo2021individually}, using adversarial formulation~\cite{grari2019fair} or constraints to reduce disparity~\cite{ravichandran2020fairxgboost, cruz2023fairgbm, barata2021fair}. In particular, FairGBM is an adaptation for gradient-boosting, which includes an update of dual variables during fitting each new weak learner to satisfy disparity fairness constraints. The results highlighted the gains in terms of computing the costs of utilizing extreme gradient boosting for fairness. No gradient boosting algorithm has yet been designed for the min-max fair formulation.

\section{Proposed Method}

We propose M²FGB, a versatile framework for subgroup fairness based on the concepts of min-max fair optimization. To achieve that, we incorporate into a loss minimization problem a term that considers the maximum group-wise loss: $\mathcal{L}(\bm{y}, f(\bm{x}))  + \max_{z} \{\overline{\mathcal{L}}_z(\bm{y}, f(\bm{x}))\}$.  $\mathcal{L}$ represents the average loss for all samples, and $\overline{\mathcal{L}}_z$ is a measure of loss for each group $z$. While the more direct approach considers that $\Lz$ is equal to $\Ls$ evaluated only with the samples from group $z$, $\Ls$ and $\Lz$ can also have different forms. This problem can be reformulated using constraints:

\begin{align}
\begin{split}
\label{eq:primal}
P: \quad \min_{f, \epsilon} \quad  (1 - \lambda) \mathcal{L}(\bm{y}, f(\bm{x})) + \lambda \epsilon \\
\textrm{s.t.}\quad \overline{\mathcal{L}}_z(\bm{y}, f(\bm{x})) \leq \epsilon \;\; \forall z \in \mathcal{Z}
\end{split}
\end{align}

Where $\lambda \in [0, 1]$ is a weight to control the strength of fairness in the model.
$\lambda = 1$ is the setting in which only the worst performance is optimized.
The min-max formulation permits the demonstration of desirable properties of fairness. 
We first present a theorem introduced by \citet{martinez2020minimax}:

\begin{theorem}[No unnecessary harm] A predictor (classifier or regressor) $f_{m}$ solution  for the min-max problem (Eq.~\ref{eq:minimax}) has all group-wise losses lower or equal than any predictor with equal group-wise loss, i.e., $f_{eq} \text{ such that }\overline{\mathcal{L}}_z(\bm{y}, f_{eq}(\bm{x})) = \overline{\mathcal{L}}_{\hat{z}}(\bm{y}, f_{eq}(\bm{x})), \forall z, \hat{z} \in \mathcal{Z}$, called ``equal predictor''.

\begin{proof}
Let's suppose by contradiction that an equal predictor $f_{eq}(x)$ has the loss from one of the groups lower than the min-max solution $f_m(x)$. Thus $\exists \hat{z}: \overline{\mathcal{L}}_{\hat{z}}(\bm{y}, f_{eq}(\bm{x})) < \overline{\mathcal{L}}_{\hat{z}}(\bm{y}, f_{m}(\bm{x}))$. Since $\max_{z \in Z} \overline{\mathcal{L}}_z(\bm{y},f_{m}(\bm{x})) \geq \overline{\mathcal{L}}_{\hat{z}}(\bm{y}, f_{m}(\bm{x})) > \overline{\mathcal{L}}_{\hat{z}}(\bm{y}, f_{eq}(\bm{x}))$ the equal predictor $f_{eq}(\bm{x})$ is also a feasible solution and, as all group-wise loss are equal, $\max_{z \in Z} \overline{\mathcal{L}}_z(\bm{y},f_{eq}(\bm{x})) < \max_{z \in Z} \overline{\mathcal{L}}_z(y,f_{m}(\bm{x}))$, what contradicts the optimality of the formulation in Equation \ref{eq:minimax}.
\end{proof}
\end{theorem}

This theorem states the min-max solution always dominates the predictor with equal losses independently of how accurate it is, i.e., each group-wise loss of the min-max solution is lower or equal to the respective loss of an equal predictor.

However, only minimizing the maximal loss might hurt the predictor's performance as reductions on other groups losses would not improve the objective function after the predictor reached the loss lower bound of the worst-performing group with the available data, which made us propose the formulation in Equation \ref{eq:primal} that keeps a weight in the loss for all samples, reducing the generalization issue for groups with small number of samples.
The parameter $\lambda$ can control how much we penalize the error in the worst-performing group in our model, as the following theorem states.

\begin{theorem}[Fairness monotonicity]\label{theo:fairness_monotone} Let's suppose that we have $\overline{\lambda}$ and $\underline{\lambda}$ such that $\overline{\lambda} > \underline{\lambda} > 0$ with respective solutions $\overline{f}(\bm{x})$ and $\underline{f}(\bm{x})$. Than, the discrepancy between the worst group-wise loss and the overall loss $\left[\max_{z} \{\overline{\mathcal{L}}_z(\bm{y}, f(\bm{x}))\} - \mathcal{L}(\bm{y}, f(\bm{x}))\right]$ for predictor $\overline{f}(\bm{x})$ is lower than predictor $\underline{f}(\bm{x})$.

\begin{proof} For simplicity, we will denote $\overline{\mathcal{L}}_z(\bm{y}, f(\bm{x}))$ as $\overline{\mathcal{L}}_z(f)$. By contradiction, assume the following: 

\begin{equation}
    \label{eq:contradiction} 
    [\max_{z} \{\overline{\mathcal{L}}_z(\overline{f})\} - \mathcal{L}(\overline{f})] > [\max_{z} \{\overline{\mathcal{L}}_z(\underline{f})\} - \mathcal{L}(\underline{f})]  
\end{equation}

By moving terms, we obtain:

\begin{equation}
    \label{eq:max_over_rel}
    [\max_{z} \{\overline{\mathcal{L}}_z(\overline{f})\} - \max_{z} \{\overline{\mathcal{L}}_z(\underline{f})\}] > [ \mathcal{L}(\overline{f}) - \mathcal{L}(\underline{f})] 
\end{equation}

\myparagraph{Case 1} $\mathcal{L}(\overline{f}) \geq \mathcal{L}(\underline{f})$. By Equation (\ref{eq:max_over_rel}) we have that $\max_{z} \{\overline{\mathcal{L}}_z(\overline{f})\} > \max_{z} \{\overline{\mathcal{L}}_z(\underline{f})\}$, what means that $\forall \lambda \in [0, 1]$: $(1-\lambda)\mathcal{L}(\overline{f})  + \lambda \max_{z} \{\overline{\mathcal{L}}_z(\overline{f}) \} \geq (1-\lambda)\mathcal{L}(\underline{f})  + \lambda \max_{z} \{\overline{\mathcal{L}}_z(\underline{f})\}$. But if $\lambda$ is set to $\overline{\lambda}$ we have a contradiction on the optimality of $\overline{f}(x)$.

\myparagraph{Case 2} $\mathcal{L}(\overline{f}) < \mathcal{L}(\underline{f})$. By the optimality of $\underline{f}$ with $\underline \lambda$ we have:

\begin{align}
    (1-\underline{\lambda})\mathcal{L}(\overline{f})  + \underline{\lambda} \max_{z} \{\overline{\mathcal{L}}_z(\overline{f})\} &\geq (1-\underline{\lambda})\mathcal{L}(\underline{f})  + \underline{\lambda} \max_{z} \{\overline{\mathcal{L}}_z(\underline{f})\} \implies \\  
    \dfrac{1 - \underline \lambda}{\underline \lambda} (\mathcal{L}(\overline{f}) - \mathcal{L}(\underline{f}))
    &\geq \max_{z} \{\overline{\mathcal{L}}_z(\underline{f})\} - \max_{z} \{\overline{\mathcal{L}}_z(\overline{f})\}
\end{align}

Similarly, we can obtain the following inequality using the optimality of $\overline f$ with $\overline \lambda$:

\begin{equation}
    \dfrac{\overline \lambda - 1}{\overline \lambda} (\mathcal{L}(\overline{f}) - \mathcal{L}(\underline{f}))
    \geq \max_{z} \{\overline{\mathcal{L}}_z(\overline{f})\} - \max_{z} \{\overline{\mathcal{L}}_z(\underline{f})\}
\end{equation}

By summing this two inequalities, we obtain that $\frac{\overline \lambda - \underline \lambda}{\overline \lambda \underline \lambda}  (\mathcal{L}(\overline{f}) - \mathcal{L}(\underline{f})) \geq 0$. As $\frac{\overline \lambda - \underline \lambda}{\overline \lambda \underline \lambda}  > 0$, we have that $ (\mathcal{L}(\overline{f}) - \mathcal{L}(\underline{f})) \geq 0$, which is a contradiction.

\end{proof}
\end{theorem}

This theorem shows that with a solution with certain gap between the worst-group loss and overall loss, we can increase the value of parameter $\lambda$ to obtain a new predictor with an equal or decreased gap. The problem defined by M²FGB (Eq.~\ref{eq:primal}) has the following Lagrangian function:

\begin{align}
\label{eq:lagrangian}
&(1 - \lambda) \mathcal{L}(\bm{y}, f(\bm{x})) + \lambda \epsilon + \sum_{z \in \mathcal{Z}} \mu_z \left(\overline{\mathcal{L}}_z(\bm{y}, f(\bm{x})) - \epsilon\right) = \nonumber \\
= &(1 - \lambda) \mathcal{L}(\bm{y}, f(\bm{x})) + \epsilon (\lambda - e^t \bm{\mu}) + \bm{\mu}^T \bm{\overline{\mathcal{L}}}
\end{align}

Where $\bm{\mu} = [\mu_1, \mu_2, \dots, \mu_{|\mathcal{Z}|}]$ and $\bm{\overline{\mathcal{L}}} = [\overline{\mathcal{L}}_{1}, \overline{\mathcal{L}}_{2}, \dots, \overline{\mathcal{L}}_{|\mathcal{Z}|}]$, $\bm{\mu}, \bm{\overline{\mathcal{L}}} \in \mathbb{R}^{|\mathcal{Z}|}$, and $e$ is the unit vector of dim $|\mathcal{Z}|$.
Dual variables $\bm{\mu} \in \mathbb{R}^{|\mathcal{Z}|}$ are incorporated by the Lagrangian function, which can be optimized as a two-player game, where one player updates $f$ to minimize the loss, while the other player updates $\mu$ to maximize the dual function~\cite{cotter2019optimization}. We employ the gradient-boosting algorithm to perform the iterative optimization. At each boosting iteration, $\bm{\mu}$ moves in the direction of the gradient, while $f$ is updated in the negative direction of the Lagrangian gradient (Eq.~\ref{eq:lagrangian}). 
At each iteration, $\epsilon$ is set to be equal to $\max_z \Lz$. We incorporate the constraint $e^T \bm{\mu} = \lambda$ in the Lagrangian function, and define the dual problem:

\begin{align*}
D: \quad \max_{\bm{\mu}} \quad &\varphi(\bm{\mu}) = \inf_f \{ (1 - \lambda) \mathcal{L}(\bm{y}, f(\bm{x})) + \lambda \epsilon  + \bm{\mu}^T( \bm{\overline{\mathcal{L}}} - \epsilon e^t) \}\\
\textrm{s.t.}\quad &e^T\bm{\mu}=\lambda, \quad\quad \bm{\mu} \geq 0.
\end{align*}

This constraint permit us to ignore $\epsilon$ while performing gradient descent. The dual variables is  $\mu$ is updated in the direction of the gradient, $\nabla \varphi(\mu) = (\overline{\mathcal{L}} - \epsilon e^t$),  and then projected to the feasible set with an Euclidean projection $\mu_p = \min_{w \in S} ||w - \mu||_2$, where $S = \{ w \in \mathbb{R}^{\mathcal{|Z|}} \mid ||w||_1 = \lambda, w \geq 0 \}$. This projection has an exact solution with a simple algorithm $O(|\mathcal{Z}|)$ (which is $<<n$) proposed by \citet{condat2016fast}.

The pseudocode code of our approach is present in Algorithm~\ref{alg:m2fgb}.

\begin{algorithm}
\caption{M²FGB}
\label{alg:m2fgb}
\SetKwInOut{Input}{Input}
\SetKwInOut{Output}{Output}
\Input{dataset $(\bm{x},\bm{y}, \bm{z})$, fair weight $\lambda$, number of iterations M, $\gamma$ learning rate, $\zeta$ dual learning rate}
\Output{model $f$}
Initialize predictor $f^0$, dual variables $\mu^0 = \lambda / |\mathcal Z|$ and $\epsilon^0 = \max_z {\Lz} (\bm{y}, f^{0}(\bm{x}))$\;
\For{$t\in \{1,\cdots,M\}$}{
$\bm{\overline{\mathcal{L}}} \gets [\overline{\mathcal{L}}_{1}(\bm{y},f^t(\bm{x})), \quad \overline{\mathcal{L}}_{2}(\bm{y},f^t(\bm{x})), \dots ]$ \; 
$\mu^{t+1} \gets \argmin_{||w||_1 = \lambda} ||w - (\mu^t + \zeta (\bm{\overline{\mathcal{L}}} - \epsilon))||_2 $ \Comment*[r]{Projected gradient ascent}
$\mathcal{G} \gets - [ (1 - \lambda) \frac{\partial \;\mathcal{L}(\bm{y}, f^t(\bm{x}))}{\partial (f(\bm{x}))} + \underset{z \in \mathcal{Z}}{\sum} \mu^t[z]  \frac{\partial \overline{\mathcal{L}}_z(\bm{y}, f^t(\bm{x}))}{\partial (f(\bm{x}))})]$ \Comment*[r]{Compute lagrangian grads}
$h^t \gets \argmin_h \sum_i[\mathcal{G}_i-h(x_i)]^2$ \Comment*[r]{Fit weak learner}
$f^{t+1} \gets f^t + \gamma h^t $ \Comment*[r]{Update ensemble model}
$\epsilon^{t+1} \gets \max_z {\Lz} (\bm{y}, f^{t+1}(\bm{x}))$ \Comment*[r]{Ensure feasibility}
}
\end{algorithm}

M²FGB needs that $\Lz$ is differentiable, which is not valid for common fairness metrics. However, similarly to the optimization of accuracy (non-differentiable) by minimizing cross-entropy (differentiable), the literature of fairness has already discussed convex and differentiable proxy functions for fairness measures. In the following section (Sec.~\ref{sec:other_fairness_metrics}) we present diverse formulations of $\Lz$. Furthermore, in Sec.~\ref{sec:experiments}, we show that, while the optimization problem optimizes a proxy function, the original measure of interest is also optimized. In Appendix~\ref{sec:convergence}, we present a study on the convergence of the proposed algorithm.

\myparagraph{Complexity} Simple gradient-boosting algorithms have complexity $O(TKd n\log n)$, where $T$ is the number of trees, $K$ the max depth of each tree, $d$ the number of features and $n$ the number of samples. M²FGB adds the complexity of the dual step: calculate the losses $\overline{\mathcal L}_z$ in time $O(n|\mathcal{Z}|)$, and update of $\mu$ in $O(|\mathcal{Z}|)$ (generally $|\mathcal Z| << 100$). For that reason, the total complexity of M²FGB is $O(TKd n\log n + T(n|\mathcal{Z}| + |\mathcal Z|))$.

\subsection{Min-Max Fairness Metrics and Proxy Functions}
\label{sec:other_fairness_metrics}

We present derivations of $\Lz$ for classification and regression tasks, including the \textbf{Equalized Loss} fairness principle for classification and regression tasks, \textbf{Equality of Opportunity} and \textbf{Demographic Parity} for classification. Under the principle of equalized loss, $\Ls$ and $\Lz$ share the same expression but are evaluated in different populations. With $\ell$ being the cross-entropy or the mean squared error, $\Ls(\bm{y}, f(\bm{x})) := \dfrac{1}{n}\sum_{i = 1}^n \ell(y_i, f(x_i))$ and $\Lz := (N_z)^{-1} \sum_{i = 1}^n \bm{1}_{[z_i = z]} \ell(y_i, f(x_i))$ where $N_z$ is the number of samples from group $z$.

When considering equality of opportunity, the loss function must only consider the performance of different groups among the true positive samples (considering that the benefit is $y = 1$). Thus, we can set a metric \textbf{TP loss}

$\Lz := (\sum_{i} \bm{1}_{[z_i = z, y_i=1]})^{-1} \sum_i \bm{1}_{[z_i = z, y_i = 1]} \ell(y_i, f(x_i))$. With the objective to maximize the WG true positive rate, we minimize the WG TP loss.

When considering demographic parity, the loss must only consider the ratio of positive predictions from each group, and this is obtained by changing the target to $1$ for every sample, defining the \textbf{P loss} $\Lz := (N_z)^{-1} \sum_i \bm{1}_{[z_i = z]} \mathcal{L}(1, f(x_i))$.  With the objective to maximize the WG positive rate, we minimize the WG P loss.

In the following sections, we will add the suffixes ``(tpr)'' and ``(eod)'' to technique names formulated for true positive rate and equal opportunity, respectively.

\section{Experiments}
\label{sec:experiments}

We present a comprehensive analysis of the M²FGB. First, we evaluate the algorithm's convergence and capability of obtaining solutions with different trade-offs between performance and fairness. Next, we also perform an intensive evaluation of the generalization of M²FGB in four benchmark datasets, employing hyperparameter optimization and evaluating different trade-off levels. Experiments will be focused on the scenario of the true positive rate and positive rate as the fairness loss and a regression task; additional results are available in Appendix~\ref{app:extra_results}. The code for the experiments will be available to support reproduction.

\myparagraph{Datasets} Experiments included four datasets commonly utilized on fairness benchmarks with different sizes. \textit{German Credit}~\cite{german} is a dataset of 1,000 individuals who took credit and were classified as bad or good risk. 4 groups were defined based on gender and age. \textit{COMPAS}~\footnote{The COMPAS dataset, a widely used benchmark in fairness studies, is included for comparative purposes. We acknowledge that algorithmic fairness is not a substitute for adressing deeper structural issues regarding incarceration and race~\cite{pruss2024prediction}.} is a software used to assess the likelihood of a defendant becoming a recidivist~\cite{angwin2016machine}. This dataset contains 6,000 assessment predictions. 4 groups were defined based on race. \textit{ACSIncome}~\cite{retiring2021ding} poses a challenge different from the previous ones by presenting more than 1 million observations. ACSIncome includes labels if individuals had an income higher than U\$50K/yearly. Race and gender were used to define 8 groups. \textit{ENEM}~\cite{alghamdi2022beyond} is a dataset with the results of Brazilian students on a national exam. Race and gender were used to define 8 groups. We used a random selection of 50,000 samples with a variation for classification and regression. A more detailed description of datasets and subgroups is available in Appendix~\ref{app:datasets}.

\myparagraph{Compared Techniques} M²FGB was evaluated against a set of techniques presented in Sec.~\ref{sec:related_works}. LGBM~\cite{guolin2017lgbm} is a baseline gradient-boosting algorithm that does not consider fairness. FairGBM~\cite{cruz2023fairgbm} is an algorithm based on gradient-boosting that optimizes for minimal disparity between groups. Both techniques were selected because they are based on gradient-boosting. MMPF~\cite{martinez2020minimax} introduced the concept of min-max fairness and utilized multi-objective optimization to find the Pareto frontier formed by group-wise losses. MinMaxFair~\cite{diana2021minimax} is also based on min-max fairness, and at each iteration, samples' weights are calculated to ensure fairness. The algorithm has a threshold for the WG loss as parameter, which can be used to set the trade-off between performance and fairness. These last two algorithms were evaluated using logistic regression as the base model fitted at each iteration. 

\begin{figure}
\centering
\includegraphics[width=0.9\textwidth]{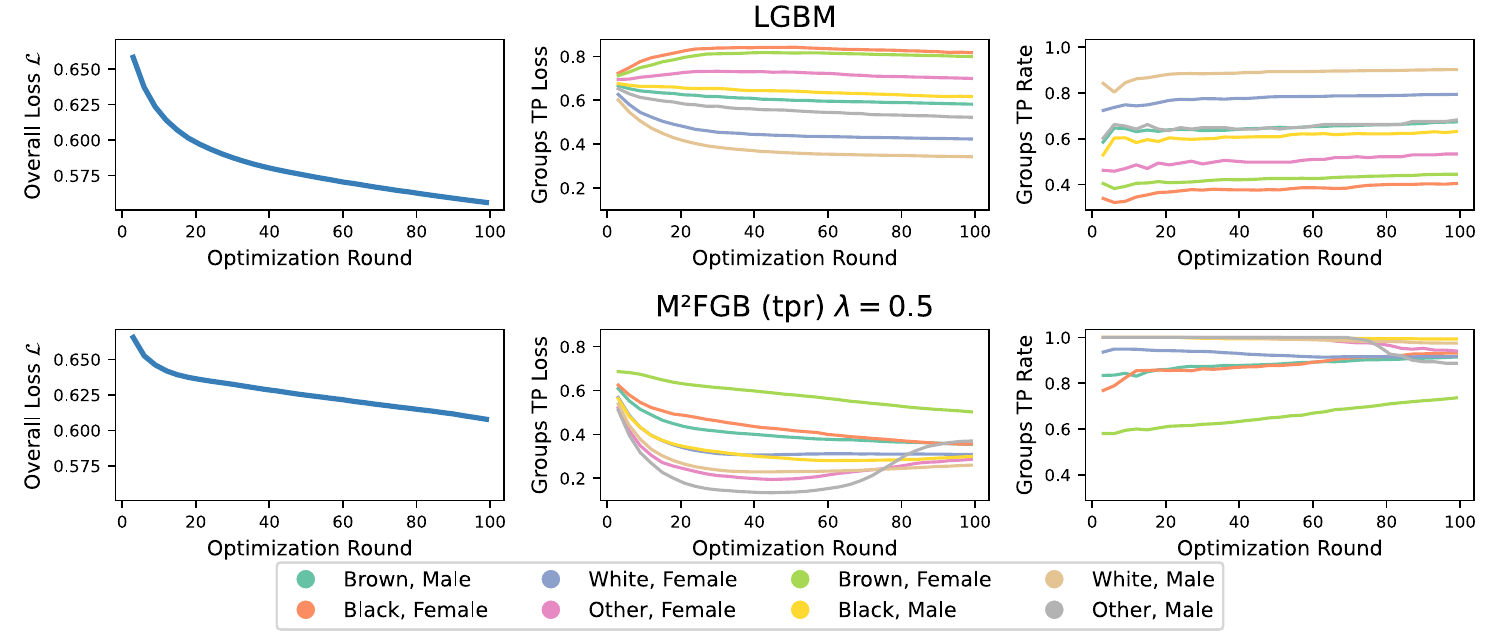}
\caption{Comparison between LGBM and M²FGB. The first column displays the decrease on log cross-entropy $\Ls$ for the model, and in the following two columns, each of the 8 lines display the evolution of the differentiable loss function ($\Lz$), the non-differentiable metric for a specific group $z$. $\lambda$ was set to 0.5.}
\label{fig:convergence_german}
\end{figure}

\subsection{Algorithm's Convergence}

To better understand the iterative optimization of M²FGB, we visualize in a illustrative example the evolution of the TP loss (as $\Lz$) and the non-differentiable metric of interest, TP rate. M²FGB optimization will increase the weights $\mu_z$ for the groups that have the worse performance, consequently, giving greater importance to reducing the loss in those samples. To study that, we compare M²FGB with LGBM that is only set to optimize for the log cross entropy. This experiment was carried out with the ENEM dataset using 8 subgroups. Both models were trained using the same hyperparameters for 100 boosting rounds. The results on training data are shown in Fig.~\ref{fig:convergence_german}. 

The final value of $\Ls$ obtained by LGBM is lower than M²FGB, which is expected, as it is the only objective of LGBM. Taking into account group losses $\Lz$, although the TP loss of some groups presented a reduction with LGBM, there is no improvement in the WG TP loss, which actually increases during optimization (red line). M²FGB increases the weight $\mu_z$ of the groups with higher TP loss during optimization, causing a significant reduction in WG TP loss (light green line). The improvement obtained by M²FGB is also observed in the TP rate of the worst group, reaching near 0.75, while it never exceeds 0.5 in the worst group with LGBM. The increase in TP loss observed in the group with gray line occurs due to the conflict between gradients from the objectives $\Ls$ and $\Lz$. Our test considered $\lambda = 0.5$, but this parameter can set the trade-off between fairness and performance, as our following experiment evaluates.

\subsection{$\lambda$ Parameter}

\begin{figure}
    \centering
  \includegraphics[width=0.9\textwidth]{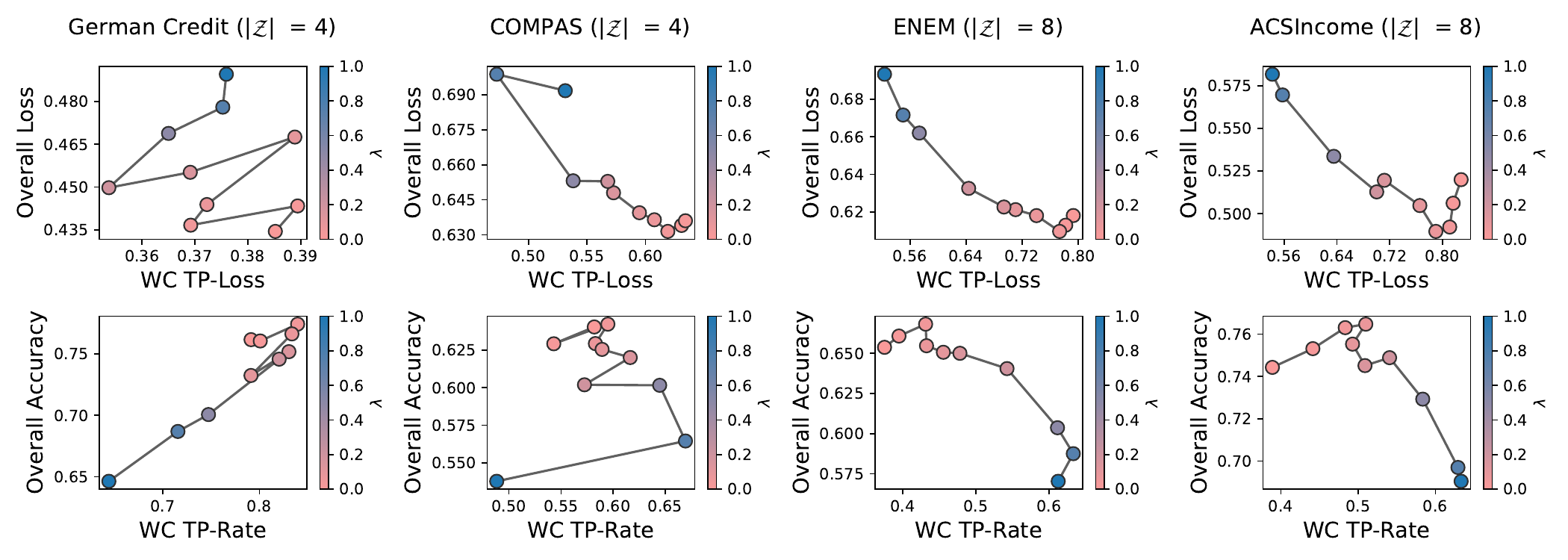}
  \caption{M²FGB solutions at different fairness strength values ($\lambda$ parameter). Increasing $\lambda$ increases performance on the worst-performing group, with a cost on the overall loss.}
  \label{fig:fair_weight}
\end{figure}

To show the usefulness of the fairness weight $\lambda$, as already analytically evaluated in Theo. \ref{theo:fairness_monotone}, we display the performance loss and fairness loss obtained with different values of $\lambda$. To do so, for each $\lambda$ we report the overall loss (log cross-entropy) and the WG TP loss on the training sample of 100 models with random hyperparameters. The random selection of hyperparameters was necessary to control for the interaction effects of $\lambda$ and the remaining parameters. The result of this procedure is shown in Fig.~\ref{fig:fair_weight}. We also included the results of the same models on the non-differentiable metrics, i.e., accuracy and WG TP rate.  Analyzing the first row of plots, increasing $\lambda$ decreases the WG TP loss (higher fairness) with an increase in the overall loss (lower performance). Taking into account the second row, increasing $\lambda$ increased the WG TP rate (higher fairness) while decreasing the accuracy (lower performance). Using $\lambda = 1$ (or values close to $1$) may hurt optimization, as each boosting round will only consider the data from the worse group, which is commonly of small sample size. This effect is more present on the smaller datasets, German and COMPAS. With moderate values of $\lambda$, information from other groups start to take part in the learning process, thus presenting a better trend. This experiment demonstrates that M²FGB finds diverse solutions according to the fairness strength utilized.

\subsection{Generalization Results on Benchmark Datasets}

Lastly, we present an extensive experiment comparing M²FGB and related techniques results on test data. We considered each technique's hyperparameters and the trade-off between performance and fairness. To do so, we employ the score utilized by \citet{cruz2023fairgbm}, which weights both objectives by a parameter $\alpha$: $(\alpha \times \text{performance} + (1 - \alpha) \times \text{fairness})$. We utilized accuracy as \textit{performance} and the WG TP rate or the positive rate as \textit{fairness} in a classification task and MSE and WG MSE in a regression task.

\myparagraph{Experimental Setting}
The data was split into 60\% train, 20\% validation, and 20\% test. Validation data was used to perform randomized hyperparameter optimization with 100 combinations. Hyperparameters included were related to fairness strength and model complexity. Due to high computational cost, MMPF and MinMaxFair with the ENEM and ACSIncome datasets were executed only with 10 iterations of the solver for logistic regression. The best model on validation was then evaluated on the testing set. The reported results are an average of 1000 repetitions. The experiments were executed on an Intel(R) Core(TM) i7-5820K CPU @ 3.30GHz with 64 GB of RAM.

\begin{figure}[t]
    \centering
    \includegraphics[width=0.9\linewidth]{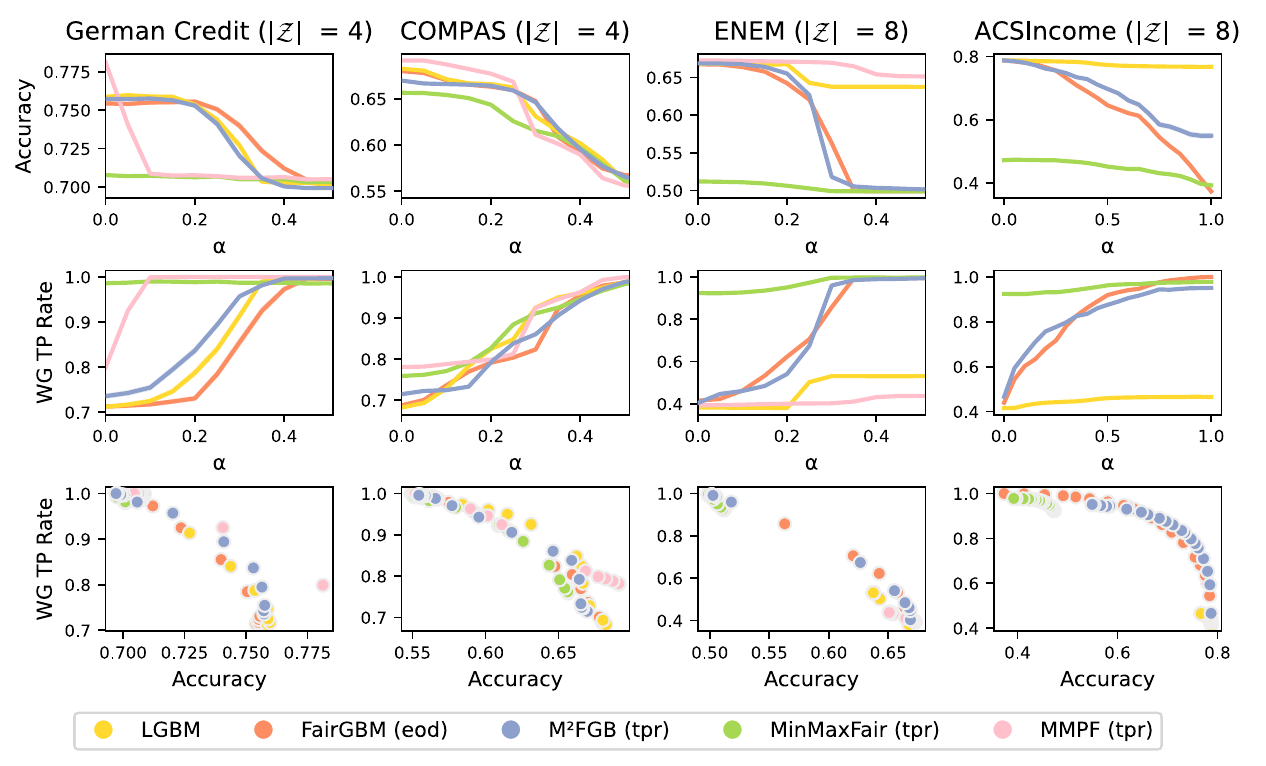}
    \caption{Average performance and fairness of algorithms in the test set of 1000 executions. 95\% confidence intervals are displayed but really tight. Hyperparameter optimization was executed at different levels in the trade-off between performance and fairness. M²FGB presented higher or competitive accuracy and WG TP rate at all studied datasets.}
    \label{fig:bench_tpr}
\end{figure}

\myparagraph{TP Rate Results} In Fig.~\ref{fig:bench_tpr} we display the results of M²FGB and compared techniques at multiple $\alpha$ values. The result of each dataset is displayed in a column. In the first three datasets, we only show $\alpha$ up to $0.5$, as higher values depicted models that predicted positive for every input. This result is observed because by only predicting the positive label, the WG TPR metric is 1, and dominates any improvement in accuracy. In general, \textbf{M²FGB} had the most steady behavior, 
achieving solutions with maximum fairness when $\alpha$ was close to $1$, and more performative solutions with lower values of $\alpha$. On the other hand, \textbf{MMPF} was not able to have good WG TP rate in ENEM, and \textbf{MinMaxFair} could not obtain models with great performance in three of the four datasets. For gradient boosting methods, \textbf{FairGBM} was able to have fair or performative models by varying $\alpha$, however, did not present better results than M²FGB in any of the datasets; and \textbf{LGBM} was unable to have a good WG TP rate in ENEM and ACSIncoime for all $\alpha$. Considering the third row of plots, we are interest in algorithms that are capable of presenting solutions at different positions in the trade-off between performance and fairness, as it was done by M²FGB in all four datasets. It is important to highlight the difference in computing costs of our proposed approach and other techniques of min-max fairness. Tab.~\ref{tab:computing_time} displays the average time to fit the best performing models ($\alpha = 0$) as a ratio of the execution time of the LGBM. Gradient-boosting algorithms, such as M²FGB, are much faster than the other approaches. FairGBM, which is not designed for min-max fairness, was faster than M²FGB; however, it has a C++ implementation.

\setlength{\tabcolsep}{1mm}
\begin{table}[t]
\centering
\begin{tabular}{lllll}
\toprule
Model & German & COMPAS & ENEM & ACSIncome \\ \midrule
LGBM  & $1$ & $1$ & $1$	 & $1$	\\
FairGBM & x0.9 & x2.2 & x0.6 & x1 \\
MMPF & x27.8 & x19.1 & x11.4 & --- \\
MinMaxFair & x15.6 & x42.1 & x11.9 & x24.0\\
M²FGB & x4.1 & x2.4 & x1.7 & x2.3 \\
\bottomrule
\end{tabular}
\caption{Relative computing cost of fair algorithms against the unfair LGBM baseline. Average results of the best-performing model. MMPF was not run with ACSIncome due to time constraints.}
\label{tab:computing_time}
\end{table}

\begin{figure}[t]
    \centering
    \includegraphics[width=0.9\linewidth]{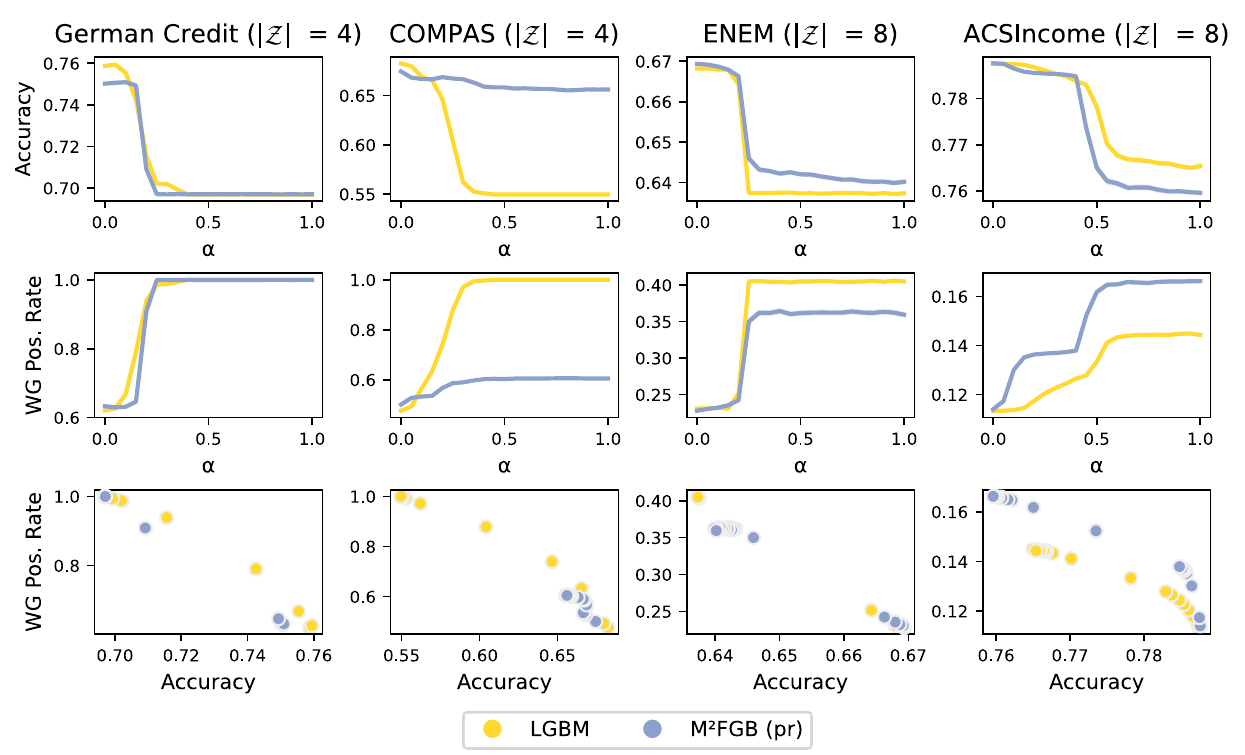}
    \caption{Average performance and fairness of hyperparam optimization performed at multiple $\alpha$ considering the positive rate criteria. This setting may result on models that only predict the positive class, as ocurred with LGBM on German Credit and COMPAS datasets, and with M²FGB on German Credit. On other datasets, M²FGB was able to obtain gains on fairness without a high cost in performance.}
    \label{fig:bench_pr}
\end{figure}

\myparagraph{Positive Rate Results}
Our approach is the only one that considers the positive rate as a measure of fairness in the min-max fairness setting. This is a valid concern as, for example, different subgroups of the population should receive the same allocation of resources in public policies. However, maximizing the minimal positive rate among groups can make it difficult to learn a strong predictor, as the model that predicts the positive class for every observation is a model that minimizes the fairness loss but is not a useful predictor. The hyperparameter optimization procedure was done considering the combined score: $(\text{WG positive rate}) \alpha + (\text{accuracy}) (1 - \alpha)$. We evaluate M²FGB against standard LGBM at Fig.~\ref{fig:bench_pr}. 

In German Credit and COMPAS, the WG pos. rate obtained by LGBM with $\alpha$ values greater than $0.2$ is exactly $1.0$, that is, the fair model learned by LGBM is a classifier that predicts the positive class for every sample. M²FGB obtained a similar result in German Credit, however, in the COMPAS dataset the proposed algorithm obtained higher values of WG pos. rate as $\alpha$ increased, without resulting in a classifier that only predicts the positive class. On ENEM and ACSIncome datasets, LGBM obtained fair models without resulting in degenerate classifiers. In particular, it reached at most 40\% of WG pos. rate on the ENEM dataset, which was higher than the result obtained by M²FGB. However, in the ACSIncome dataset, M²FGB resulted in models that better explored the performance-fairness trade-off, reaching a WG pos. rate of 16\%, which was not reached by LGBM. M²FGB obtains competitive results in diverse points in the fairness and performance trade-off; however, it poses greater advantages with larger datasets, such as ACSIncome.

\begin{figure}[t]
    \centering
    \includegraphics[width=\linewidth]{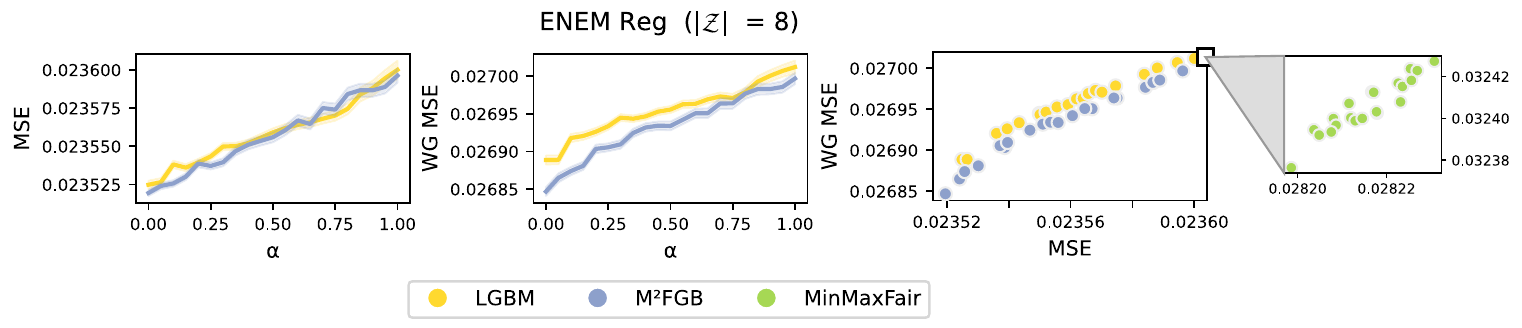}
    \caption{Average performance and fairness of hyperparam optimization performed at multiple $\alpha$ in a regression task.}
    \label{fig:bench_mse}
\end{figure}

\myparagraph{Regression Results} A regression experiment was performed using the ENEM dataset, with the target $y$ being the fraction of correct answers. The result is displayed in the last column of Fig.~\ref{fig:bench_mse}. We evaluated only M²FGB, LGBM, and MinMaxFair, as the remaining techniques can not be easily altered for regression. In this study, the lower the overall MSE and WG MSE, the better. MinMaxFair performance was worse than LGBM and M²FGB and was omitted in the first two rows, being highlighted in the separated plot. In this experiment, increasing $\alpha$ only resulted in worse results for the WG MSE metric. This might have occurred due to models not being able to generalize on a small subgroup of the test set. In the scatter, it is possible to see that M²FGB obtained the best solution both in terms of performance and fairness. This result on regression highlight the flexibility of M²FGB capabilities, being suited for diverse tasks.

\section{Discussion and Conclusion}

Our experiments showed that M²FGB can achieve competitive performance and fairness values across multiple datasets in improved computing time. This section discusses some of its limitations and concludes the paper.

\myparagraph{Fairness Metrics} M²FGB can work with several learning losses and fairness metrics, and we present some variations. However, finding differentiable proxy functions for the evaluated metrics was necessary to incorporate such fairness metrics. We based our formulations on the log cross-entropy, already utilized as a proxy function for accuracy. FairGBM has already used proxy functions for fairness metrics~\cite{cruz2023fairgbm}.

\myparagraph{Overfit} The objective function is optimized within the boosting rounds, permitting us to learn a strong predictor that minimizes WG loss. At each weak learner added to the model, gradient boosting algorithms reduce the loss of the training set, which can lead to overfitting when the number of iterations is large. In that way, an increased number of iterations of M²FGB, which can reduce fairness, can also result in a model that will not generalize out-of-sample. It is important to tune M²FGB hyperparameters, allowing it to learn more rounds with simple weak learners.

We proposed M²FGB, a general framework for min-max subgroup fairness models that can benefit from gradient-boosting optimization algorithms.
Our method is general and can incorporate different convex predictive losses and fairness metrics, with variations of fairness metrics presented.
By focusing on minimizing the error of the worst-performing subgroup, our approach increases fairness without penalizing the remaining subgroups heavily, as discussed theoretically and presented in experiments.

\section{Acknowledgements}

This project was supported by the brazilian Ministry of Science, Technology and Innovations, with resources from Law nº 8,248, of October 23, 1991, within the scope of PPI-SOFTEX, coordinated by Softex and published Arquitetura Cognitiva (Phase 3), DOU 01245.003479/2024-10.

\newpage

\appendix

\section{Analysis of M²FGB Convergence}
\label{sec:convergence}

In this section, we analyze the convergence of M²FGB by showing that the algorithm finds a saddle-point solution of the primal-dual problems. To do so, we assume that $\Ls, \Lz$ are convex and $K$-smooth. As each iteration of the proposed algorithm ensures that $(f^t, \epsilon^t, \mu^t)$ are feasible, we need to evaluate the duality gap of a solution $f, \epsilon, \mu$:

\begin{equation}
\label{eq:duality_gap}
G(f, \epsilon, \mu) = L(f, \epsilon, \mu) - \min_{f', \epsilon'} L(f', \epsilon', \mu)
\end{equation}

where $L$ is the Lagrangian function:

\begin{equation}
L(f, \epsilon, \mu) = (1 - \lambda) \Ls(f) + \lambda \epsilon + \sum_{z \in \mathcal{Z}} \mu_z(\Lz(f) - \epsilon)
\end{equation}

Due to the extra constraint included by M²FGB ($\sum_z \mu_z = \lambda$), the Lagrangian function can be simplified to $L(f, \mu) = (1 - \lambda) \Ls(f) + \sum_{z \in \mathcal{Z}} \mu_z\Lz(f)$, without the dependence on $\epsilon$. $G$ measures how far the primal solution $(f)$ is from the optimal solution when $\mu$ is fixed. We have that $G(f, \epsilon, \mu) > 0$ for any solution, and if each iteration of M²FGB decreases the value, we can conclude that it converges. We are interested in the measure:

\begin{equation}
\begin{split}
G(f^{t+1}, \mu^{t+1}) - G(f^{t}, \mu^{t}) = \left[ L(f^{t+1}, \mu^{t+1}) - L(f^t, \mu^t) \right] - \left[  \min_{f, } L(f', \mu^{t+1}) -  \min_{f'} L(f', \mu^{t}) \right] = \\
= \left[ \left(L(f^{t+1}, \mu^{t+1}) - L(f^{t}, \mu^{t+1})\right) + \left(L(f^{t}, \mu^{t+1}) - L(f^t, \mu^t)\right) \right] - \left[  \min_{f, } L(f', \mu^{t+1}) -  \min_{f'} L(f', \mu^{t}) \right]
\end{split}
\end{equation}

We study each of these three terms. First, we consider the update $f^t \to f^{t+1}$. Using the fact that $\Ls, \Lz$ are $K$-smooth, we can use the upper bound on the second-order Taylor approximation of $L$ at point $(f^t, \mu^{t+1})$. Then, evaluate it on the point $(f^{t+1}, \mu^{t+1})$. This gives us:

\begin{equation}
L(f^{t+1}, \mu^{t+1}) - L(f^{t}, \mu^{t+1}) \leq - \gamma \langle \nabla_f L(f^t, \mu^{t+1}) , h^t\rangle + \dfrac{K \gamma^2}{2}||h^t||^2
\end{equation}

This difference can be ensured to be negative if the learning rate is small enough, satisfying:

\begin{equation}
    \gamma \leq \dfrac{-2 \langle \nabla_f L(f^t, \mu^{t+1}), h^t \rangle }{K ||h^t||^2}
\end{equation}

As $h^t$ is optimized to approximate $-\nabla_f L = -\left( (1-\lambda) \nabla \Ls + \lambda \sum_z \mu_z \Lz \right)$, it is only necessary that $h^t$ is aligned enough with $-\nabla_f L(f^t, \mu^{t+1})$ to satisfy $\langle \nabla_f L(f^t, \mu^{t+1}), h^t \rangle < 0$, and consequently the right-hand side of this bound is positive. We now consider the update $\mu^t \to \mu^{t+1}$:

\begin{equation}
    L(f^{t}, \mu^{t+1}) - L(f^t, \mu^t) = \sum_z (\mu_z^{t+1} - \mu_z^{t}) \Lz (f) \leq ||\mu^{t+1} - \mu^{t}|| \cdot ||\bm{\overline {\mathcal{L}}} (f^t)||
\end{equation}

Where the last step used the Cauchy-Schwarz inequality. The euclidean projection $\textrm{Proj}_S(\cdot)$ into a convex set $S$ is 1-Lipchitz, and by definition, if $S = \{ v \in \mathbb{R}^{|\mathcal Z|} \mid || v ||_1 = \lambda, v \geq 0\}$, then $\mu^{t+1} = \textrm{Proj}_S(\mu^{t} + \zeta (\bm{\overline{\mathcal{L}}}(f^t) - \epsilon^t)) $ and $\mu^{t} =  \textrm{Proj}_S(\mu^{t})$ (because $\mu^t$ is already in $S$). Then, using the fact that $\textrm{Proj}_S(\cdot)$ is 1-Lipchitz:

\begin{equation}
\begin{split}
    L(f^{t}, \mu^{t+1}) - L(f^t, \mu^t) &\leq ||\mu^{t+1} - \mu^{t}|| \cdot  ||\bm{\overline {\mathcal{L}}} (f^t)|| \\
    & \leq ||\mu^{t} + \zeta (\bm{\overline{\mathcal{L}}}(f^t) - \epsilon^t) - \mu^t || \cdot  ||\bm{\overline {\mathcal{L}}} (f^t)|| \\
    &= \zeta ||(\bm{\overline{\mathcal{L}}}(f^t) - \epsilon^t) || \cdot   ||\bm{\overline {\mathcal{L}}} (f^t)||
\end{split}
\end{equation}

Lastly, we consider the gap between optimal primal solutions for the Lagrangian with $\mu^t$ and $\mu^{t+1}$. Let $f^\star$ be the solution of $\min_f L(f, \mu^{t})$. We have than that:

\begin{equation}
    \min_{f, } L(f', \mu^{t+1}) -  \min_{f'} L(f', \mu^{t}) \leq L(f^\star, \mu^{t+1}) - L(f^\star, \mu^{t}) \leq \zeta ||(\bm{\overline{\mathcal{L}}} (f^\star) - \epsilon^t) || \cdot  ||\bm{\overline{\mathcal{L}}} (f^\star)||
\end{equation}

Where the last step is done by using again the Cauchy-Schwartz inequality and the fact that the euclidean projection is 1-Lipchitz. Finally, we have that the duality gap is:

\begin{equation}
\begin{split}
G(f^{t+1}, \mu^{t+1}) - G(f^{t}, \mu^{t}) =&  \left[ \left(L(f^{t+1}, \mu^{t+1}) - L(f^{t}, \mu^{t+1})\right) + \left(L(f^{t}, \mu^{t+1}) - L(f^t, \mu^t)\right) \right] - \left[  \min_{f, } L(f', \mu^{t+1}) -  \min_{f'} L(f', \mu^{t}) \right] =\\
=& - \gamma \langle \nabla_f L(f^t), h^t \rangle + O(\gamma^2 + \zeta)
\end{split}
\end{equation}

Which can be make negative by selecting learning rates $\gamma$ and $\zeta$.

\section{Description of Datasets and Subgroups}
\label{app:datasets}
\begin{figure}
    \centering
    \includegraphics[width=0.7\linewidth]{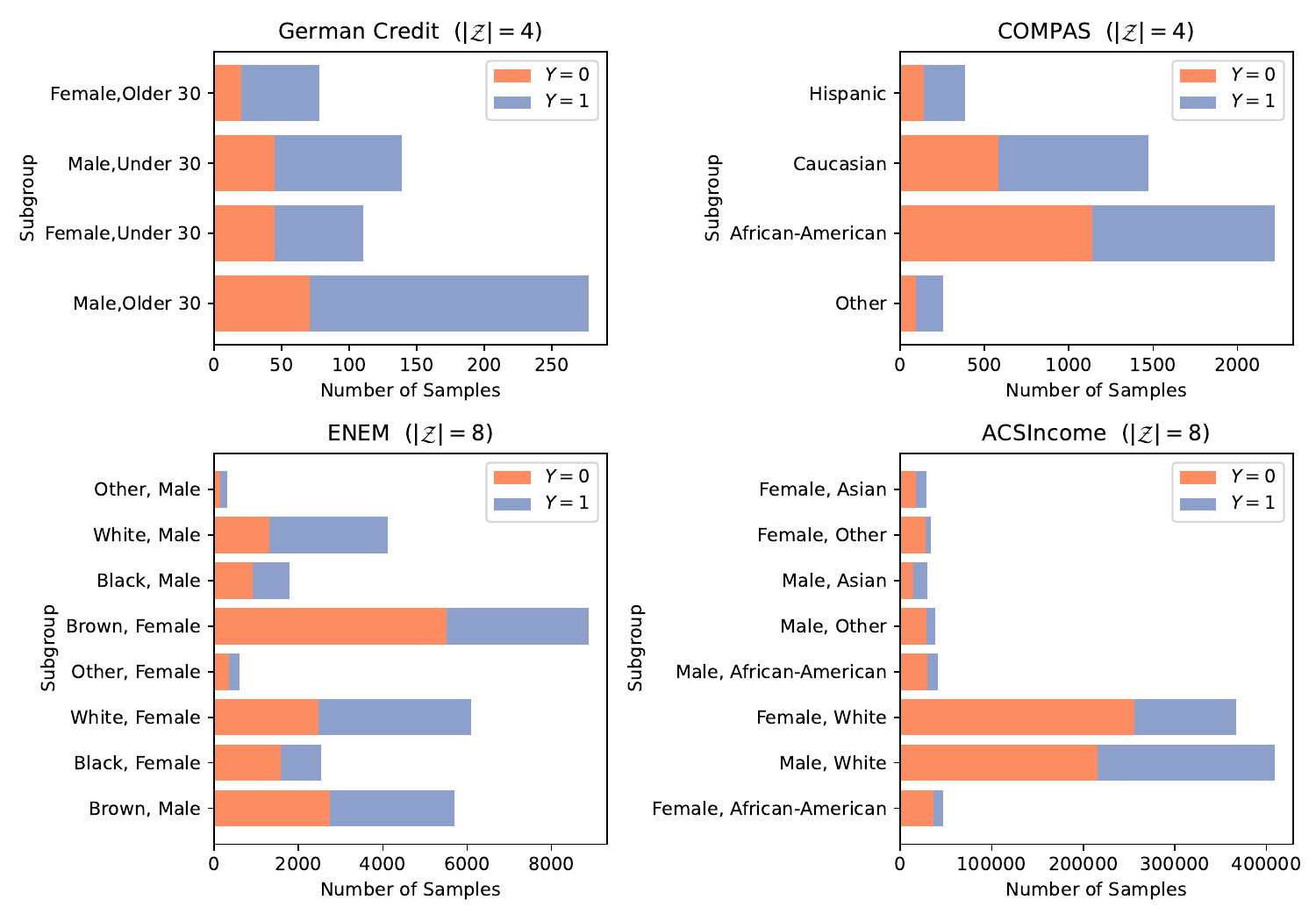}
    \caption{Datasets and subgroups utilized in our experiments, with the distribution of labels $Y$ among each group.}
    \label{fig:prop_subgroups}
\end{figure}

The datasets considered in our experiments are common in the fairness literature; however, most works focus on analyzing binary groups. For defining 4 and 8 subgroups, we used the intersectionality of different sensitive attributes. Fig.~\ref{fig:prop_subgroups} presents the distribution of samples in each subgroup for all datasets and also the distribution of the labels $y$. All datasets have between 1\% and 13\% of samples in the smallest group. All datasets were preprocessed with one hot encoding of categorical variables and standard scaling of numerical ones. We present some extra details of datasets and subgroups:

\begin{itemize}
    \item German Credit contains information on clients from a bank in Germany, with a classification of them in good or bad payers, with 70\% of payers being classified as good ones. The subgroups were defined with the intersection of gender and 2 age groups --- if the client was older than 30 years old. The smallest group is \textit{(Female, Older 30)} with 78 (13\%) samples in the training set.
    \item COMPAS dataset was obtained from the study by \citet{angwin2016machine}. Their work studied the predictions of COMPAS, a software utilized in US courts for assessing the likelihood of recidivism among defendants. Their study highlighted the discrimination of the software towards African Americans, predicting higher likelihoods for individuals of this race. We considered four groups defined by race, with the smallest one, \textit{(Other)}, having 256 (6\%) of samples in the training set.
    \item ENEM (\textit{Exame Nacional do Ensino Médio}) dataset compromises information about high school students in a national exam about four areas of study. From each student, it is collected a set of personal information regarding their socioeconomic conditions. This dataset was introduced in fairness studies by \citet{alghamdi2022beyond}, and we utilized the same pre-processing as theirs, using 50,000 random samples and the target as the grading in human sciences discretized in two classes. All features in this dataset are categorical and were one-hot encoded. While \citet{alghamdi2022beyond} considered two groups defined by gender, we utilized the intersection of race and gender, i.e., $\{\text{White, Black, Brown, Asian and Native (Others)}\} \times \{\text{Male, Female}\}$. The smallest group was \textit{(Other, Male)} with 299 (1\%) samples in the training set. We also generated a variation of the dataset --- a dataset for a regression task with 50,000 samples, in which the grading was not discretized in two classes but a score between $0$ and $1$.
    \item ACSIncome~\cite{retiring2021ding} is a big dataset from a family of datasets proposed to replace the commonly utilized dataset Adult. The particular dataset utilized has census information regarding 1,664,500 US citizens with the label if the yearly income was higher than U\$50,000. The subgroups were defined as the intersection between gender $\{\text{Male, Famale}\}$ and race $\{\text{White, African-American, Asian, Others}\}$. The smallest group was \textit{(Female, Asian)} with 28,599 (3\%) samples in the training set.
    
\end{itemize}

\section{Experiments Details}

For all methodologies, we considered reasonable hyperparam spaces for performing random hyperparam tuning. The hyperparam spaces were defined based on the details present in the original papers. The hyperparam space needs to be wide due to the great diversity of datasets, including the number of observations, the number of features, and feature types. See the supplemental code for a complete description of hyperparam spaces. Tab.~\ref{tab:methods_hyperparams} presents the hyperparameters considered for each algorithm. 

Train, validation, and test splits were performed with a stratified approach for each group. That is, samples of each group were separated into the train, validation, and test sets with an equal proportion of labels in all sets. Then, these sets were combined in the final training, validation, and test data. This was necessary to ensure that each group was observed in all data splits and because the smallest groups tend to have more unbalanced labels than the majority ones.

To avoid 1,000 executions of random hyperparameter optimization, 1,000 (500 when considering ACSIncome) combinations of hyperparameters were sampled, the models trained, and saved onto disk to create a pool of models. Then, to perform a trial of random hyperparameter optimization, we sample 100 models from the pool of previously trained models and select the best-performing one on the validation set to be evaluated on the test set. A similar methodology was employed by \citet{cruz2023fairgbm}.

Small adjustments were necessary to the original implementation of MinMaxFair by \citet{diana2021minimax}. First, to run the regression experiment, the standard linear model was replaced with a Ridge Regression (linear regression with L2 regularization). Similar to the classification scenario, SAGA solver with limited iterations was executed to ensure model training in a feasible computing time.

\begin{table}[]
\centering
\begin{tabular}{ll}
\toprule
Algorithm & Hyperparameters \\ \midrule
LGBM & \makecell[cl]{n. of estimators, learning rate, n. of leaves, \\ min. hess. sum in child, L2 weight} \\ \hline
M²FGB & \makecell[cl]{Same parameters as LGBM, \\ fairness weight and multiplier learning rate} \\ \hline
FairGBM & \makecell[cl]{Same parameters as LGBM \\ constraint tol. and multiplier learning rate} \\ \hline
MMPF & \makecell[cl]{n. of iterations, $\alpha$, $\mathcal{K}_{\min}$, L2 weight} \\ \hline
MinMaxFair & \makecell[cl]{n. of iterations, $a$ and $b$ parameters, \\  $\gamma$ fairness weight, L2 weight} \\ \bottomrule
\end{tabular}
\caption{Hyperparameters considered for optimization of M²FGB and compared algorithms.}
\label{tab:methods_hyperparams}
\end{table}

\section{Extra Results}
\label{app:extra_results}

We present extra results considering the equalized loss metric in the classification setting.

\begin{figure}
    \centering
    \includegraphics[width=\linewidth]{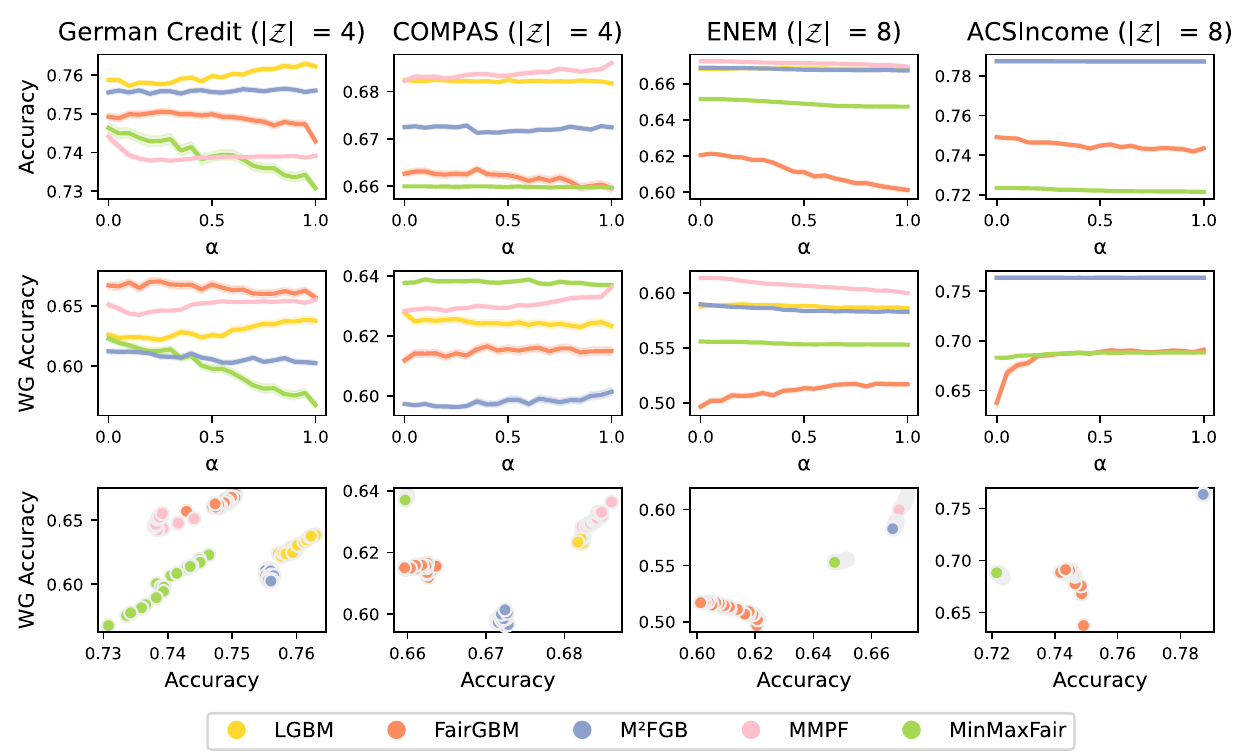}
    \caption{Average performance and fairness of hyperparam optimization performed at multiple $\alpha$ considering the equalized loss criteria.}
    \label{fig:bench_eq}
\end{figure}

\myparagraph{Equalized Loss Results} Fig.~\ref{fig:bench_eq} presents the results of algorithms according to the equalized loss criteria. To perform hyperparameter optimization, the combined score of $(\text{WG accuracy}) \alpha + (\text{accuray}) (1 - \alpha)$ was used. During the execution of experiments, we identified that the small number of samples in the smallest group of each dataset caused the hyperparameter optimization procedure to overfit the validation data, as the WG accuracy is only calculated from a group of the population (commonly the smallest). This overfit on the validation data was a challenge in obtaining good fairness scores on the test data. In that sense, M²FG and LGBM presented similar performance and fairness scores, showing that due to the high performative capabilities of LGBM (which M²FGB also presents), reducing the overall loss without fairness regularization was highly capable of improving min-max fairness.

We analyze the results for each dataset. In German Credit, FairGBM had the WG accuracy at any $\alpha$ level, however, this result was not present on other datasets. M²FGB, similar to LGBM, reached a high accuracy in this dataset, yet it was not capable of reaching higher values of WG accuracy. On the COMPAS dataset, the highest fairness was reached by MinMaxFair, followed by MMPF. M²FGB obtained good accuracy results, however, its fairness metric was lower than that of the compared algorithms. In the ENEM and ACSIncome datasets, FairGBM and MinMaxFair had the lowest values of performance and fairness against the compared algorithms. MMPF obtained the highest WG accuracy in ENEM, also presenting an accuracy competitive with M²FGB, however, it presents a high computational cost and was not executed in ACSIncome. M²FGB and LGBM presented similar results in ACSIncome, greatly exceeding other algorithms in performance and fairness.

\end{document}